\documentclass[conference]{IEEEtran}
\IEEEoverridecommandlockouts
\usepackage{cite}
\usepackage{amsmath,amssymb,amsfonts}
\usepackage{algorithmic}
\usepackage{graphicx}
\usepackage{textcomp}
\usepackage{xcolor}
\def\BibTeX{{\rm B\kern-.05em{\sc i\kern-.025em b}\kern-.08em
    T\kern-.1667em\lower.7ex\hbox{E}\kern-.125emX}}
\begin{document}

\title{Quantifying Nematodes through Images: Datasets, Models, and Baselines of Deep Learning\\
\thanks{This work is supported by Biotechnology and Biological Sciences Research Council with project number BB/X01200X/1.}
}

\author{\IEEEauthorblockN{Zhipeng Yuan}
\IEEEauthorblockA{\textit{\small Department of Computer Science} \\
\textit{University of Sheffield}\\
Sheffield, The United Kingdom \\
zhipeng.yuan@sheffield.ac.uk}
\and
\IEEEauthorblockN{Nasamu Musa}
\IEEEauthorblockA{\textit{\small Soils, Crops and Water} \\
\textit{RSK ADAS Ltd High Mowthorpe}\\
Malton, The United Kingdom \\
nasamu.musa@adas.co.uk}
\and
\IEEEauthorblockN{Katarzyna Dybal}
\IEEEauthorblockA{\textit{\small Agriculture and Environment Department} \\
\textit{Harper Adams University}\\
Newport, The United Kingdom \\
kdybal@harper-adams.ac.uk}
\and
\IEEEauthorblockN{Matthew Back}
\IEEEauthorblockA{\textit{\small Agriculture and Environment Department} \\
\textit{Harper Adams University}\\
Newport, The United Kingdom \\
mback@harper-adams.ac.uk}
\and
\IEEEauthorblockN{Daniel Leybourne}
\IEEEauthorblockA{\textit{\small Department of Evolution, Ecology and Behaviour} \\
\textit{University of Liverpool}\\
Liverpool, The United Kingdom \\
Daniel.Leybourne@liverpool.ac.uk}
\and
\IEEEauthorblockN{Po Yang}
\IEEEauthorblockA{\textit{\small Department of Computer Science} \\
\textit{University of Sheffield}\\
Sheffield, United Kingdom \\
po.yang@sheffield.ac.uk}
}

\maketitle

\begin{abstract}
Every year, plant parasitic nematodes, one of the major groups of plant pathogens, cause a significant loss of crops worldwide. 
To mitigate crop yield losses caused by nematodes, an efficient nematode monitoring method is essential for plant and crop disease management.
In other respects, efficient nematode detection contributes to medical research and drug discovery, as nematodes are model organisms.
With the rapid development of computer technology, computer vision techniques provide a feasible solution for quantifying nematodes or nematode infections.
In this paper, we survey and categorise the studies and available datasets on nematode detection through deep-learning models.
To stimulate progress in related research, this survey presents the potential state-of-the-art object detection models, training techniques, optimisation techniques, and evaluation metrics for deep learning beginners.
Moreover, seven state-of-the-art object detection models are validated on three public datasets and the AgriNema dataset for plant parasitic nematodes to construct a baseline for nematode detection.
\end{abstract}

\begin{IEEEkeywords}
Deep learning, Object detection, Nematode detection, Image analysis
\end{IEEEkeywords}

\section{Introduction}
Plant parasitic nematodes are one of the major groups of plant pathogens causing crop and plant diseases \cite{savary2019global}, which seriously impact food security and production losses.
Specifically, the parasitic of nematodes accounts for over 10\% of annual crop losses and cost roughly 100 billion U.S. dollars worldwide \cite{abad2008genome} \cite{yuan2023pestdss}. 
This means that timely plant-parasitic nematode detection is necessary to address the increased food demand caused by population growth estimated \cite{dubois2011state}.

To meet these challenges, a series of studies have been proposed to detect vermiform adult nematodes and cysts (deceased female nematodes carrying eggs) of nematodes \cite{eves2015metagenetic}. 
Techniques of nematode detection have been categorised as biochemical detection and molecular diagnostics \cite{shao2023current}.
Specifically, biochemical detection methods include isoenzyme analysis \cite{carneiro2017methods} and mass spectrometry \cite{vega2018improvement} to identify the nematode species by protein fingerprints of different nematode species.
In contrast, molecular diagnostics focus on the DNA characteristics of different species of nematodes, which are detected by PCR technologies \cite{sikder2020novel} or isothermal amplification technologies \cite{ahuja2021diagnosis}.
Although some studies \cite{shao2023recombinase} have been devoted to providing time-saving and efficient sampling and detection methods for nematode detection in the field, there is still a gap between aforementioned detection methods and low-cost nematode detection methods.

The recent employment of computer technology stimulates the development of pest and disease management in precision agriculture and provides new solutions for low-cost nematode detection methods based on image analysis.
In particular, the ability of deep learning models to automatically extract visual features facilitates the recognition of tiny objects in complex image backgrounds \cite{yuan2022lightweight}.
Therefore, deep learning models are able to detect nematodes from microscope images \cite{bogale2020nematode} or images from high magnification cameras \cite{kranse2022low}.
Compared to the aforementioned methods and traditional morphological visual detection methods, deep learning models do not require extensive substantial expertise and training for end users and avoid subjective judgement \cite{yuan2023automatic}.
Although deep learning technology is gradually being applied to nematode detection tasks, there is still a lack of accurate and robust detection models due to the need for annotated training data and the challenge of designing optimal model structures for nematode detection tasks.

As a consequence, there is an urgent need for more efforts to develop low-cost nematode detection models, especially for cysts detection in the early stage to mitigate yield losses caused by nematodes.
To facilitate research on deep learning-based nematode detection models, a comprehensive survey of deep learning-based object detection models with corresponding baselines for the nematode detection task is necessary.

The main contribution of this review is to construct a comprehensive baseline for state-of-the-art object detection models in the nematode detection tasks to facilitate research on deep learning-based nematode detection solutions.
The baseline covers the performance of seven state-of-the-art object detection models on three public nematode datasets and a dataset constructed for crop parasitic nematode detection.
Meanwhile, this review not only presents a survey of relevant studies, but also classifies and discusses feasible deep learning-based detection models, training tricks, evaluation metrics, and available datasets for the nematode detection task in order to guide deep learning beginners in exploring nematode detection tasks.
In addition, a discussion of challenges and model optimisation directions for the nematode detection task is attached to expose the flaws of current research.


\section{Methodology}\label{sec:method}
The methodological design for this review includes research questions, search strategy, and study selection criteria.
The definition of research questions describes the need and motivation for this review. 
Therefore, the development of the search strategy and selection criteria is influenced by research questions.

Compared to other techniques, deep learning models have presented unexpected accuracy in computer vision tasks due to their ability to learn visual features automatically. 
Therefore, exploring the feasibility of vision-based deep neural networks on the nematode detection task has drawn the attention of researchers.  
The main research question that the review is concerned with and attempts to answer is "\textit{How can deep neural networks solve the challenge of low-cost nematode detection?}".
Moreover, a set of secondary research questions are defined to answer the main research question, namely,
\begin{itemize}
\item SQ1 - What deep learning methods have been used in the nematode detection task?
\item SQ2 - What are the available datasets?
\item SQ3 - What are the state-of-the-art object detection models?
\item SQ4 - What are the methods for model training and optimisation?
\item SQ5 - What are the metrics for model evaluation?
\end{itemize}

The search strategy focuses on digital scientific databases including Google Scholar, IEEE Xplore Digital Libray, ACM Library,  Elsevier Scopus, and Springer Link. 
Meanwhile, search keywords are defined as the following generic string,
("deep learning" OR "deep neural network" OR "convolutional neural network" OR "object detection") AND ("nematode detection").
Whilst the above search keywords are able to identify research relevant to nematode detection, the secondary questions about potential object detection models are not adequately presented by the search results.
Therefore, the second search keywords are defined as follows, 
("deep learning" OR "deep neural network" OR "convolutional neural network") AND ("object detection").

Since deep learning is a relatively new technique, studies from the previous ten years (from 2014 to 2023) are selected in the research results.
This work uses different selection strategies for the search results of the two search keywords.
For the search results of nematode detection studies, this review focuses on whether these studies describe the methods and results clearly.
Meanwhile, the studies on object detection models are selected through the reproducibility of the work, especially whether the code and data are available.
Therefore, the number of forks and stars of the corresponding code repository is the main quantification metric.
Methods, where the sum of forks and stars does not exceed three thousand, are filtered.

\section{Deep Learning in Nematode Detection}\label{sec:results}

To the best of our knowledge, this is the first review that focuses exclusively on nematode and cysts detection in agriculture through deep-learning technology. 
Related to the nematode detection, 19 loop-mediated isothermal amplification assays \cite{ahuja2021diagnosis} were investigated, while the relationship between nematode and crop and management practices were discussed \cite{zhang2020fungi}.
In addition, surveys focusing on specific crops \cite{abd2019plant}\cite{abd2020biological} or nematode species\cite{bairwa2017techniques} were proposed.

In this section, we present the review results for the studies of nematode detection and potential object detection technologies that contribute to answering the five corresponding secondary questions in this review.

\subsection{SQ1 - Deep Learning in Nematode Detection}

\begin{table*}[tbp]
\caption{Details of articles for nematode detection}
\begin{center}
\begin{tabular}{ccccccc}
\hline
\textbf{Article} & \textbf{Type of Detection} & \textbf{Models/Algorithms} & \textbf{Task type} & \textbf{Type of Input} & \textbf{Num. of Classification} & \textbf{Evaluation Metrics}\\
\hline 
 \cite{qin2023deep} & Indirect Detection & YOLOv5 & Object Detection & Multispectral; Visible & 2 & mAP; Detection Speed \\
\hline
\cite{qing2022nemarec} & Direct Detection & CNN & Classification & Microscope & 19 & Accuracy \\
\hline
 \cite{zhu2021algorithm} & Direct Detection & Attention-UNet & Segmentation & Microscope &  - & Accuracy \\
\hline
 \cite{lu4213402nema} & Direct Detection & Xception model & Classification & Microscope & 40 & Accuracy \\
\hline
 \cite{shabrina2023comparative} & Direct Detection & CNN & Classification & Microscope & 11 & F1; Precision; Recall \\
\hline
 \cite{fudickar2021mask} & Direct Detection & Mask R-CNN &  ISegmentation & Microscope & 1 & Precision \\
\hline
 \cite{wang2022detection} & Indirect Detection & YOLOv5-CMS &  Object Detection & Visible & 1 & mAP\\
\hline
 \cite{zhang2022sem} &Direct Detection & SEM-RCNN& Object Detection & Microscope & 21 & mAP \\
\hline
\cite{jimenez2019segnema} & Direct Detection & SegNema & Segmentation & Microscope & 13 & Accuracy \\
\hline
 \cite{niu2020low} & Indirect Detection & DNN; SVM; RF; DT& Classification & Reflectance & 4 & Accuracy  \\
\hline
 \cite{zhang2022counting} & Indirect Detection & VDNet  & Segmentation & Visible & 1 & Accuracy \\
\hline
 \cite{chen2020cnn} & Direct Detection & CNN; Canny & ISegmentation & Microscope & 1 & Precision; Recall \\
\hline
\cite{banerjee2023cnn} & Indirect Detection & CNN; SVM & Classification & Visible & 4 & Accuracy; F1 \\
\hline
\cite{oliveira2019analysis} & Indirect Detection & UNet & Segmentation &  Visible & 3 & Precision; Recall; F1 \\
\hline
\cite{nakasi2021poster} & Direct Detection & GoogleNet;AlexNet & Classification & Microscope & 3 & AUC \\
\hline
\cite{lin2020using} & Direct Detection & CNN & Regression & Microscope & - & Accuracy \\
\hline
\cite{pham2022classification} & Direct Detection & LSTM & Classification  & Time series & - & Accuracy\\
\hline
\cite{faster2023natalia} & Direct Detection & Faster RCNN & Object Detection & Microscope & 2 & Accuracy \\
\hline
\cite{akintayo2018deep} & Direct Detection & CNN & Object Detection & Microscope & 2 & Accuracy \\
\hline
\cite{abade2022nemanet} & Direct Detection & CNN & Classification & Microscope & 5 & Accuracy \\
\hline
\multicolumn{7}{l}{Visible: RGB images; mAP: Mean Average Precision; CNN: Convolutional Neural Network; F1: F1 scores; ISegmentation: Instance Segmentation;   } \\
\multicolumn{7}{l}{DNN: Deep Neural Network; SVM: Support vector machines; RF: Random Forest; DT: Decision Tree; Canny: Canny Edge Detector;}\\
\multicolumn{7}{l}{AUC: Area under the receiver operating characteristic curve; LSTM: Long short-term memory}
\end{tabular}
\label{tab1}
\end{center}
\end{table*}

To answer this question, this work searches 163 publications from 5 data sources and filters out literature reviews \cite{arjoune2022soybean}, non-deep learning methods \cite{chen2019instance}, and non-nematode detection studies \cite{rani2023pathogen} based on the selection strategy.
Due to the need for data for deep learning and the difficulty of obtaining annotations, 19 studies are extracted as effective deep learning-based nematode detection methods.
The investigated studies are divided into two groups, namely, direct and indirect detection of nematodes. 
Direct detection methods rely on nematode morphological characteristics to classify \cite{qing2022nemarec} or count \cite{zhang2022sem} nematodes directly from microscope \cite{fudickar2021mask} or camera images \cite{kranse2022low} of nematode samples.
Indirect nematode detection methods analyse the crop leaves \cite{oliveira2019analysis} and infected parts \cite{wang2022detection} of nematodes by multispectral \cite{qin2023deep} or camera \cite{zhang2022counting} images. 

The direct detection methods translate the nematode detection task into different computer tasks including image classification \cite{qing2022nemarec}, object detection \cite{faster2023natalia}, segmentation \cite{jimenez2019segnema}, and instance segmentation \cite{chen2020cnn}, depending on the purpose of the study.
The image classification models focus on assigning corresponding labels to individual image inputs.
Therefore, a web application integrated a CNN \cite{qing2022nemarec} provided five feeding type labels through a local microscope image, such as the head or tail of a nematode.
Meanwhile, a comparative analysis of CNN \cite{shabrina2023comparative} for nematode classification was done under a similar setup as the above work. 
The optimal model, CoAtNet-0, outperformed the other 14 models and obtained an average type accuracy of 97.86\% over an F1 score of 0.9803 \cite{shabrina2023comparative}.
Object detection models draw bounding boxes and assign labels to objects of interest in one input image.
For example, Faster RCNN was used to detect plant-parasitic and non-parasitic nematodes from microscope images and achieved 87.5\% accuracy\cite{faster2023natalia}.
Moreover, SEM-RCNN \cite{zhang2022sem} was proposed to detect multiple classes of microorganisms under environmental microorganisms.
Compared to the object detection task, the segmentation task draws the edges of objects of interest more accurately. 
In other words, the segmentation model assigns a label to each pixel in the image to classify the image region.
A modified UNet model was used to segment nematodes in digital microscopy images \cite{jimenez2019segnema} with nearly three thousand manually labelled images.
Attention mask \cite{zhu2021algorithm} was used to improve the nematode segmentation accuracy of UNet in microscopy images. 
However, a limitation of segmentation models is the inability to distinguish between instances of the same categorised object, especially when instances overlap or are connected.
Therefore, instance segmentation models are proposed to address this challenge.
For example, Mask RCNN \cite{fudickar2021mask} combined an object detection model with a segmentation model to assign pixel-level taxonomic labels and mark individual instances of the \textit{Caenorhabditis elegans (C. elegans)} nematode under low magnification microscopy.
Due to the visual characteristics of nematode objects, a study \cite{chen2020cnn} detected the skeletal position of potato cyst nematodes by CNN and mapped nematode segmentation templates in microscope images by a Canny edge detector.

For indirect detection, most research \cite{qin2023deep}\cite{liu2023observer}\cite{oliveira2019analysis} was devoted to detecting nematode infection or plant diseases caused by nematodes at large scales through drone or aerial imagery.
For example, the effectiveness of six different classification algorithms, including Neural Networks, Support Vector Machine, Random Forest, AdaBoost, Nearest Neighbors, and Decision Tree for detecting nematode infection levels by the reflectance of radio frequencies of leaves was evaluated in a case study on walnuts \cite{niu2020low} where nematode infection levels were categorised into four levels.
A modified VGG-16 network, VDNet \cite{zhang2022counting}, achieved 86\% accuracy in detecting pint wilt trees caused by nematodes via drone imagery.
Similar work was done on coffee crops to detect nematode-infected crops by the UNet model \cite{oliveira2019analysis}.
With the development of multispectral sensors, multispectral images were utilised in the detection of pine wilt disease and 98.7\% mAP was obtained with the assistance of YOLOv5 model\cite{qin2023deep}.
In addition, a study \cite{banerjee2023cnn} for plant diseases in citrus crops used a model structure combining CNN and SVM to classify nine crop diseases by individual plant leaves.
A study of root-knot nematodes (\textit{Meloidogyne}) detection in cucumber \cite{wang2022detection} detected sites of nematode infection and assess disease severity for selection of resistant cucumber varieties through a modified YOLOv5 and RGB images collected from an experimental environment.

\begin{figure*}[tbp]
\centerline{\includegraphics[width=0.8\textwidth]{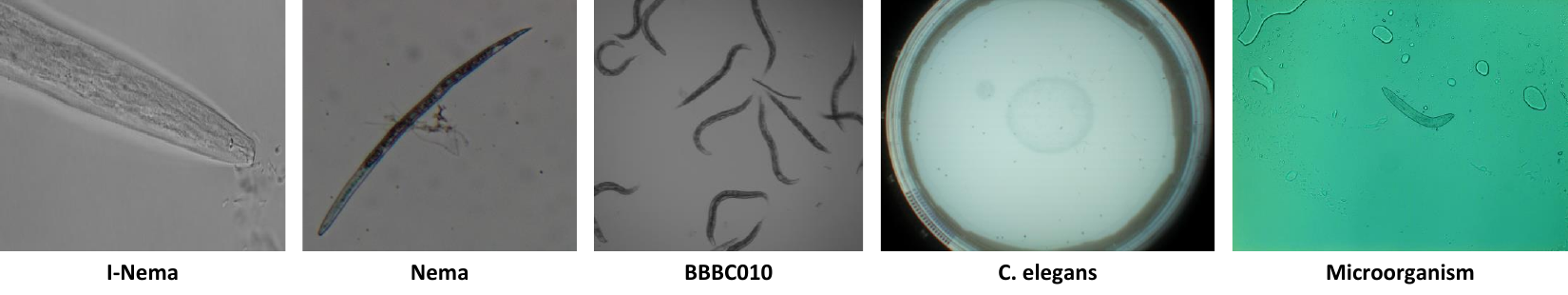}}
\caption{Examples from public datasets.}
\label{fig:dataset}
\end{figure*}

Several studies have explored the feasibility of predicting nematode developmental stages \cite{lin2020using} and nematode activity \cite{pham2022classification} to assist with the studies of plant infection resistance genes and drug discovery. 
Specifically, CNN was utilised to accurately measure the physiological age of \textit{C. elegans} on the scale of days and achieved 88.92\% accuracy for anti-ageing drug screening and genetic screening studies \cite{lin2020using}.
The study of automatic recognition of nematode behavioural patterns \cite{pham2022classification} extracted four interpretable features from a correlation matrix describing nematode shape to define nematode motivation and achieved 73.49\% accuracy in a five classification task through LSTM to assist with the study of neurological function, genetic variation, and motor sensation.

In addition to deep learning-based methods, some traditional methods of computer vision \cite{pun2022detection}, such as energy-based segmentation \cite{chen2019instance} and superpixel segmentation \cite{chen2022learning} have also been used in nematode detection tasks to alleviate the data dependency of deep learning.

\subsection{SQ2 - Public Datasets}

Constructing datasets that have the same data distribution as real-world application scenarios is crucial for training deep neural networks.
However, there are limited publicly available nematode detection datasets identified in the literature survey due to the cost of data collection and data labelling.
The largest available public dataset \cite{lu4213402nema} of microscopic nematode images provided over 9,000 images for classification tasks. 
The images showed details of nematodes at large magnifications.
In contrast, the \textit{C. elegans} nematode dataset \cite{fudickar2021mask} provided complete images under the microscope.
Table~\ref{tab:dataset} and Figure~\ref{fig:dataset} show the details of these datasets.

\begin{table}[tbp]
\caption{Details of public image datasets for nematodes }
\begin{center}
\begin{tabular}{cccc}
\hline
\textbf{Dataset} & \textbf{Images} & \textbf{Types} & \textbf{Classification} \\
\hline
I-Nema\cite{lu4213402nema} & 9,215 & Classification & 40 \\
\hline
Nema\cite{abade2022nemanet} & 3,063 & Classification & 5 \\
\hline
\textit{C. elegans}\cite{fudickar2021mask} & 1,908 & ISegmentation & 1 \\
\hline
BBBC010\cite{ljosa2012annotated} & 100 & ISegmentation & 2 \\
\hline
Microorganism\cite{Sabban_SinfNet_Microorganism_image_2023} & 498 & ISegmentation & 1 \\
\hline
\end{tabular}
\label{tab:dataset}
\end{center}
\end{table}

\section{Object Detection Model}

\subsection{SQ3 - Object Detection Models Based on Deep Learning}

\begin{figure*}[tbp]
\centerline{\includegraphics[width=0.8\textwidth]{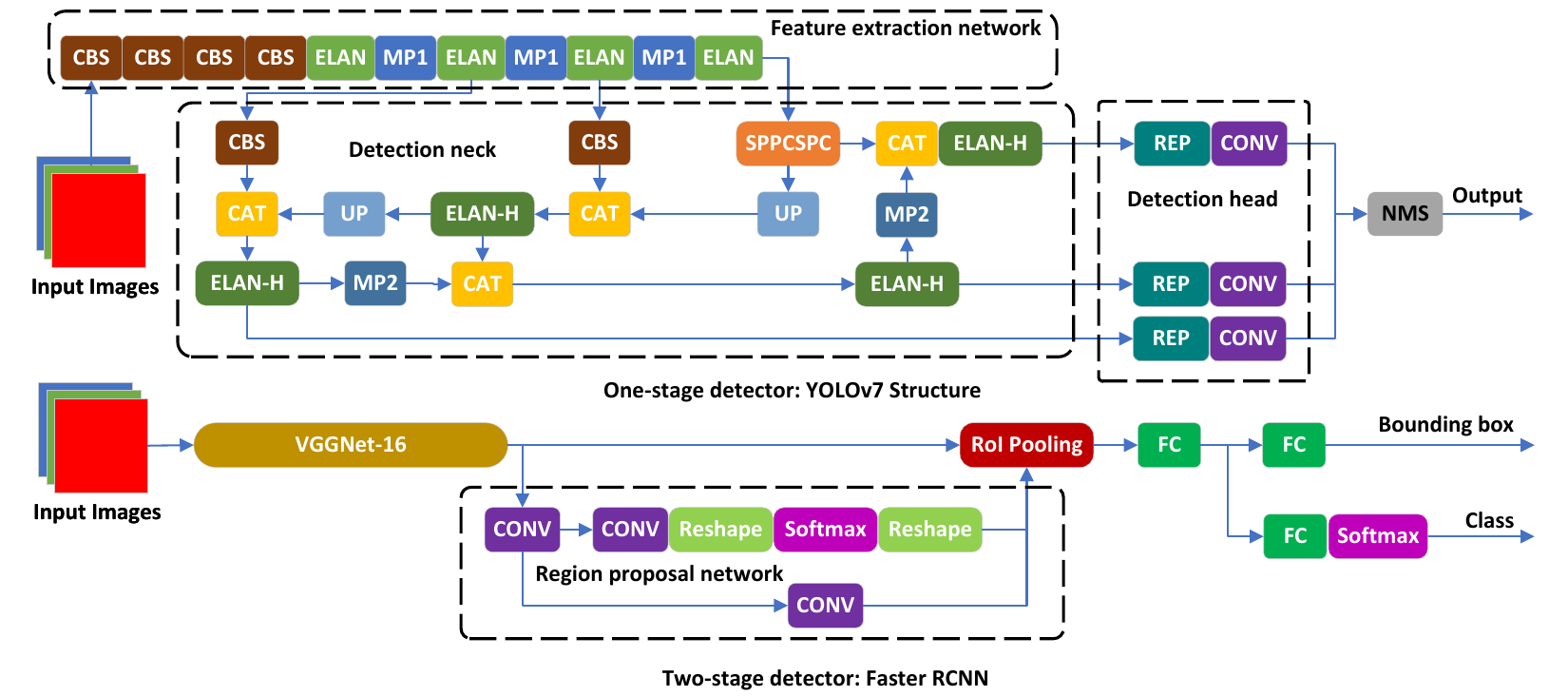}}
\caption{Model Structures of Object Detection Model. CBS, ELAN, MP1, MP2, ELAN-H, and SPPCSPC are the composite model structure in YOLOv7 \cite{wang2022yolov7}. CONV, NMS, Reshape, and FC are convolutional network modules with pooling modules, non-maximum suppression, reshape modules, and full connection modules, respectively. }
\label{fig:model_strucuters}
\end{figure*}

In the survey of current nematode detection research, four computer vision tasks are identified, including classification, object detection, segmentation, and instance segmentation.
Among them, object detection and instance segmentation models have the potential in achieving the quantification of nematodes in images.
However, since instance segmentation models require additional algorithms to calculate the number of different instances and rely on object detection models from a model design perspective, this survey focuses on object detection models.

Object detection is a classical task in the field of computer vision therefore a large number of relevant literature reviews are identified during literature search processing.
For example, a literature review \cite{10028728} comprehensively surveyed over 20 years of research on object detection methods, covering important milestone detectors, detection datasets, evaluation metrics, and optimisation methods.
Meanwhile, another literature survey \cite{liu2020deep} focused on the description of the methodology covering more than 300 works and summarised the detection framework, object feature representation, proposal generation, etc.

According to the structures of the object detection model and their functions, the popular deep learning-based object detection models are categorised as one-stage stage and two-stage detectors.
The significant difference between the two types of models is the presence or absence of a structure to generate proposals in the middle of the inference process, as shown in Figure~\ref{fig:model_strucuters}.
The proposal refers to bounding boxes that need to be further optimised.

Models with proposal-generating structures are referred to as two-stage detectors.
Popular two-stage detectors include RCNN\cite{girshick2014rich}, Fast RCNN\cite{girshick2015fast}, and Faster RCNN\cite{ren2016faster}.
The region-based convolutional neural network, RCNN\cite{girshick2014rich}, is an early form of object detection model based on deep learning, incorporating some of the structure and methods of traditional object detection methods, such as selective search for region proposals and SVM for classification.
Since CNNs need to process thousands of region proposals in RCNN, the detection efficiency is lower compared to current detection models.
The subsequently proposed Fast RCNN\cite{girshick2015fast} and Faster RCNN\cite{ren2016faster} are dedicated to solving this problem.
Fast RCNN\cite{girshick2015fast} used RoI projection to map the feature maps of the neural network onto the input image, thus reducing repetitive computation and using fully connected layers instead of SVMs.
Faster RCNN\cite{ren2016faster} combined with the region proposal network to generate region proposals achieved faster detection and 42.7\% mAP on the coco dataset.
Faster RCNN was considered the most accurate object detection model for a long time.

In contrast, common one-stage models have the advantage of faster detection speed and lighter model size including the YOLO\cite{wang2022yolov7} model family, CenterNet\cite{duan2019centernet}, and DETR \cite{carion2020endtoend}.
With a large number of training techniques and better model substructures proposed, the YOLOv5 \cite{glenn_jocher_2022_7347926} achieved similar accuracy to Faster RCNN.
The one-stage model does not require any structure for generating regional proposals.
The three types of one-stage models have different structures.
In addition to data pre-processing and post-processing, the models of YOLO have three components, including the feature extraction network, the neck and the detection head. 
Advances in the YOLO family of models focus on the optimisation of model substructures such as decoupled heads \cite{ge2021yolox} and ELAN structures \cite{wang2022yolov7}.
CenterNet\cite{duan2019centernet} did not have a neck structure and achieved an average precision of 45.1\% on the COCO dataset \cite{lin2015microsoft}.
In contrast to the aforementioned models, DETR \cite{carion2020endtoend} was a transformer-based object detection model that solved the problem of hyperparameters setting for pre-processing and post-processing.
However, DETR used a large number of model parameters in exchange for improved accuracy.
As a result, YOLOv5 remains the most popular model in GitHub according to the number of stars.
Currently, one-stage object detection models are still active research areas and new studies are constantly being proposed, such as YOLO-NAS using network structure search.

\subsection{SQ4 - Model Training and Optimisation Technologies}
In addition to the structure of the object detection model, the training and optimisation techniques of the model have a significant impact on the performance of the object detection model.
Common training techniques and optimisation techniques include loss functions, data augmentation, training auxiliary head, model quantisation, and model reparameterisation.

The loss function is a necessary component for training deep learning models. 
The loss functions are defined differently according to the definition of post-processing in object detection models.
This review summarises a generic form of loss functions for object detection models as
\begin{equation}
Loss(p,p^*,t,t^*)=L_{cls}(p,p^*)+\beta L_{box}(t,t^*), \label{eq:loss}
\end{equation}
where $L_{cls}$, $L_{box}$, $p$, $p^*$, $t$, and $t^*$ are the loss function for classification, loss function for bounding boxes, prediction of classification, ground truth of classification, prediction of bounding boxes, and ground truth of bounding box.
The classification loss functions in the object detection tasks are similar to the loss functions in the classification task using mean square error \cite{redmon2016you}, L2 loss, or cross-entropy loss \cite{ren2016faster} to measure differences in probability distributions.
In addition, focal loss is able to solve the problem of data imbalance.
For classification loss, the difference between object detection tasks and classification tasks lies in the assignment of classification probabilities.
The predicted bounding box without objects of interest is defined as negative samples assigning 0 as classification probability.
The presence of an object is measured by the intersection-over-union (IoU) between prediction and ground truth.
Meanwhile, IoU and its variants were used in many studies to compute the confidence in the presence of objects \cite{glenn_jocher_2022_7347926}.
The loss function for bounding boxes is used to optimise the position and size of the bounding box.
In addition to the above IoU with its variants, the L1 \cite{girshick2014rich} and L2 \cite{redmon2016you} loss functions were used to optimise the bounding box with respect to four variables including centre location, width, and height.

Data augmentation is widely used in the field of deep learning to compensate for data deficiencies and enhance model robustness by extending training datasets with synthetic data.
In computer vision tasks, data augmentation methods express invariants in the task, such as scale invariance and rotational invariance.
For object detection tasks, the annotations of bounding boxes are adjusted following the data augmentation methods. 
Widely used data augmentation methods include image flipping, image cropping, image rotation, image translation, colour space transformations, and mosaics.
The colour space transformations consist of global adjustments to transparency, saturation, hue and value.
Meanwhile, the implementations of Mosaic have a variety of methods including merging multiple images with or without overlapping and merging multiple images with transparency adjusting or cropping.

Meanwhile, the auxiliary head as a new deep supervision technique improves the accuracy of the model by adding auxiliary loss guidance during the model training phase.
Specifically, coarse-to-fine auxiliary head supervision was used in the study of YOLOv7 \cite{wang2022yolov7} to balance the number of positive and negative samples in the object detection task.
Earlier studies of object detection models were just labelled with bounding boxes of ground truth, so there were a large number of negative sample bounding boxes in an image.
To alleviate this problem, the auxiliary head with the same structure as the detection head is added to the model during model training.
For assigning samples, more positive samples are assigned as coarse labels to the auxiliary head based on the IoU for ground truth, which reduces the difficulty of models to learn the object features.

In addition to the improvement of accuracy, object detection speed and model size have been the focus of attention in the research.
Model quantisation and reparameterisation \cite{wang2022yolov7} for CNN are two effective ways to reduce model size in addition to model structure optimisation.
Model quantisation reduces memory usage, hard disk usage, and computational consumption by converting the float-point type of model parameters to smaller types of model parameters.
In the studies of object detection, post-training quantization \cite{ge2021yolox} was used to obtain a lighter model at the cost of accuracy.
In addition, model reparameterisation for CNN relies on the mathematically equivalent properties of convolutional computation.
In other words, the jump-link structure that is widely used in CNNs can be equivalently converted into a simple convolution operation during computation.
Therefore, YOLOv7 \cite{wang2022yolov7} considers reparameterisation in the model structure design stage and transforms the complex structure of the model into a simple chained CNN after training to obtain a lighter model with a faster detection speed.

\subsection{SQ5 - Evaluation Metrics}
The positive and negative samples of the detection results need to be defined for evaluating the performance of object detection models.
The common definition uses the IoU value between the predicted bounding box and the ground truth to determine positive and negative samples.
The predicted bounding boxes with a large IoU value are defined as positive samples marking most of the area of the object of interest.
A common threshold is 0.5.
After defining positive and negative samples, the precision and recall of object detection models are defined as 
\begin{align}
Precision(c)=\frac{TP_c}{TP_c+FP_c}, \label{eq:precision} \\
Recall(c)=\frac{TP_c}{TP_c+FN_c}, \label{eq:precision} 
\end{align}

where $TP_c$, $FP_c$, and $FN_c$ are the number of true positive samples, false positive samples, and false negative samples for class $c$, respectively.
The mean average precision (mAP) is defined as the average area under the Precision-Recall curve,
\begin{equation}
mAP=\frac{1}{C}\sum_{c\in C}\int_{0}^{1} Precision(c) dRecall(c), \label{eq:mAP}
\end{equation}
where $C$ is the set of categories in the task.
In addition to the detection accuracy, the size of the model and the speed of detection as the performance are evaluated by the number of parameters and Frames per second (FPS).

\section{Experiments and baseline}\label{sec:experiments}
While a number of object detection models have been applied to nematode detection tasks, these efforts lack comparison with baseline models due to the rapid growth in the field of object detection models.
Therefore, this work compares seven object detection models on four nematode datasets.

\subsection{Experiment Setup}

\begin{figure}[tbp]
\centerline{\includegraphics[width=0.4\textwidth]{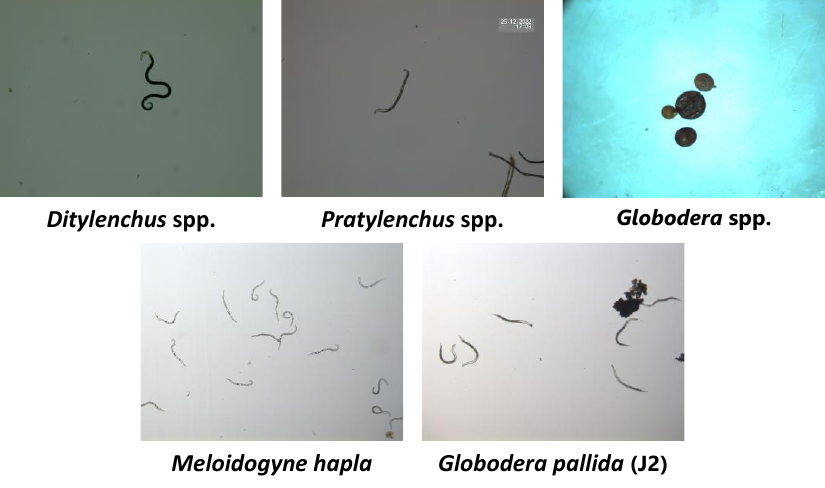}}
\caption{Examples from AgriNema Dataset.}
\label{fig:data_samples}
\end{figure}

In this study, the mAP with an IoU threshold of 0.5 is used to evaluate the accuracy of object detection models.
Meanwhile, the validated object detection models use the same training settings as the original work to improve the reproducibility of this work, which means the training tricks including data augmentation and auxiliary head are used during training.
This baseline is constructed for the lightweight model.
Therefore, the smallest model structure settings from the original works are used in our work, such as Faster RCNN with VGG \cite{ren2016faster}, YOLOv7 tiny \cite{wang2022yolov7}, and DETR with ResNet50 \cite{carion2020endtoend}.
All the models are trained in 400 epochs with pre-trained weights.

The annotations of public instance segmentation nematode datasets are converted to object detection annotations.
In addition to public datasets, this work constructs a nematode detection dataset, AgriNema, to stimulate the application of nematode detection in agriculture.
The AgriNema dataset including 525 images collected from microscopy is used to evaluate object detection models for \textit{Meloidogyne hapla}, \textit{Globodera pallida}, \textit{Pratylenchus}, \textit{Ditylenchus}, and cyst of nematodes.
Figure~\ref{fig:data_samples} presents the images from the AgriNema dataset.
All the datasets for evaluation are partitioned into a training set, validation set, and testing set in the ratio of 8:1:1.

\subsection{Baseline for Nematode Detection}

Table~\ref{tab:model_performance} presents the mAP0.5 of 7 object detection models on 4 nematode detection datasets, where YOLOv6 outperform other models on AgriNema with 96.53\% mAP. 
The performance of the models varies on different datasets due to model and dataset characteristics.
Specifically, the performance of Faster RCNN drops significantly on the tiny object C.elegans dataset due to the lack of a pyramid aggregation network for fusing multi-layer features.
The DETR model as an end-to-end object detection model avoids the object detection post-processing setting at the cost of model weights.
Therefore, DETR has no accuracy advantage, especially for the Microorganism dataset with a more complex image background using different devices for image collection.
The significant difference between the YOLO series models lies in the model backbone network and a large number of training techniques.
In addition to the impact of model structure, data preprocessing also affects the performance of YOLOv7.
\begin{table}[tbp]
\caption{Mean Average Precision of Object detection models}
\begin{center}
\begin{tabular}{ccccc}
\hline
\textbf{Models} & \textbf{AgriNema} & \textbf{C. elegans} & \textbf{BBBC010} & \textbf{Micro.} \\
\hline
Faster RCNN & 81.27\% & 19.8\% & 79.97\% & 86.29\% \\
\hline
YOLOv5 & 95.6\% & \textbf{93.1\%} & 85.6\%  & \textbf{90.4\%}\\
\hline
YOLOX & 53.75\% & 91.26\% & 36.54\% & 75.8\% \\
\hline
YOLOv6 & \textbf{96.53\%} & 84.75\% & \textbf{90.31\%} & 89.27\% \\
\hline
YOLOv7 & 53.9\% & 35.7\% & 79.7\% & 22.2\% \\
\hline
YOLOv8 & 94.5\% & 84.9\% & 89\%  & 88.1\% \\
\hline
DETR & 86.1\% & 56.1\% & 72.3\% & 16.1\% \\
\hline
\multicolumn{5}{l}{\textit{C. elegans}: \textit{C. elegans} dataset\cite{fudickar2021mask}; BBBC010: BBBC010 dataset\cite{ljosa2012annotated};}\\
\multicolumn{5}{l}{ Micro.: Microorganism\cite{Sabban_SinfNet_Microorganism_image_2023}}
\end{tabular}
\label{tab:model_performance}
\end{center}
\end{table}

\section{Discussions and Conclusions}\label{sec:conclusion}

In the last decade, deep learning-based object detection models have made impressive progress and have been applied in the field of nematode detection.
This survey not only reviews the work using deep learning to detect nematodes through images and available datasets in the last 10 years, but also exposes potential object detection models, training, and optimisation techniques by presenting the widely noticed work in the field of object detection.
Moreover, the AgriNema dataset is constructed for the study of nematode detection methods in agriculture. 
Meanwhile, a baseline is constructed including the performance of 7 state-of-the-art object detection models on 4 microscope image datasets to stimulate progress in related research.

Based on the literature survey and the construction of a baseline of nematode detection methods, a set of challenges for the research of deep learning-based nematode detection are identified.
For the nematode detection task, the primary challenge is to propose a low-cost detection solution in the fields.
The solution includes detection methods, software systems, sampling methods, and equipment for sampling to enable nematode detection in the field.
While deep learning provides relatively low-loss detection methods, sampling methods remain a challenge.

For the development of deep learning-based nematode detection, the major factor limiting relevant research at present is the lack of large-scale available data, for example, datasets from samples of soil extracts. 
The development of zero-shot learning or few-show learning techniques has the potential to mitigate this challenge.
Meanwhile, lightweight detection models are essential for the field to reduce the need for equipment. 
The aforementioned model quantisation and reparameterisation are necessary model optimisation methods.
Based on the visual characteristics of nematodes, the nematode detection task is categorised as tiny object detection. 
Therefore, optimisation for tiny object detection is beneficial to provide a more efficient solution for nematode detection.
Finally, a generic challenge for deep learning is the lack of interpretability, which hinders the application of methods in industry.
Thus exposing the reasons for the classification of the object of interest facilitates the review of model results.

\bibliographystyle{ieeetr}
\bibliography{references}


\end{document}